\documentclass[final]{cvpr}
\PassOptionsToPackage{prologue,table}{xcolor}
\usepackage{xcolor}
\definecolor{cvprgray}{RGB}{242,242,242}
\usepackage{times}
\usepackage{epsfig}
\usepackage{graphicx}
\usepackage{amsmath}
\usepackage{amssymb}
\usepackage{comment}
\usepackage{lipsum}

\usepackage[pagebackref,breaklinks,colorlinks]{hyperref}

\begin{document}

\title{Mask to reconstruct: Cooperative Semantics Completion for Video-text Retrieval}


\author{Han Fang\thanks{Equal contributions.} \and Zhifei Yang${^*}$ \and Xianghao Zang \and Chao Ban\and Hao Sun
}

\maketitle

\begin{abstract}

Recently, masked video modeling has been widely explored and significantly improved the model's understanding ability of visual regions at a local level. 
However, existing methods usually adopt random masking and follow the same reconstruction paradigm to complete the masked regions, which do not leverage the correlations between cross-modal content. 
In this paper, we present \textbf{MA}sk for \textbf{S}emantics \textbf{CO}mple\textbf{T}ion (\textit{MASCOT}) based on semantic-based masked modeling. 
Specifically, after applying attention-based video masking to generate high-informed and low-informed masks, we propose Informed Semantics Completion to recover masked semantics information. 
The recovery mechanism is achieved by aligning the masked content with the unmasked visual regions and corresponding textual context, which makes the model capture more text-related details at a patch level.
Additionally, we shift the emphasis of reconstruction from irrelevant backgrounds to discriminative parts to ignore regions with low-informed masks. 
Furthermore, we design dual-mask co-learning to incorporate video cues under different masks and learn more aligned video representation. 
Our \textit{MASCOT} performs state-of-the-art performance on four major text-video retrieval benchmarks, including MSR-VTT, LSMDC, ActivityNet, and DiDeMo. Extensive ablation studies demonstrate the effectiveness of the proposed schemes.



\end{abstract}

\section{Introduction}
Video-text retrieval is a fundamental task in multi-modal understanding at the video-text level \cite{luo2020univilm}. This process involves searching for a returned video or captions with a given cross-model query and has gained increasing attention from researchers \cite{fang2022transferring,chen2020fine,wu2021hanet,luo2022clip4clip, dzabraev2021mdmmt}. In the past years, several video-text benchmarks \cite{caba2015activitynet,xu2016msr,chen2011collecting,rohrbach2017movie,anne2017localizing} have been proposed to measure performance, which advances the development of video-text retrieval \cite{liu2022ts2,xue2022clip, gorti2022x,liu2023revisiting}.

Traditional video-text retrieval methods typically utilize fixed expert networks \cite{dzabraev2021mdmmt,faghri2017vse++,liu2019use} to extract spatial, motion, and other features for multimodal fusion \cite{yu2018joint}. Recently, advancements in image-text pre-trained models \cite{RN26,jia2021scaling} have shown remarkable generalization in video-text understanding \cite{gao2021clip,zhu2022pointclip}. Despite their success, pre-training in the video-text domain requires substantial resources for collecting large amounts of annotated video-text pairs \cite{fang2022transferring}. Consequently, there is a growing interest in transferring knowledge from pre-trained vision-language models to video-text retrieval tasks \cite{ma2022x,luo2022clip4clip,zhao2022centerclip}. Several temporal modeling methods \cite{liu2022ts2,liu2023revisiting} have been proposed for integration with CLIP \cite{RN26} to enhance global video representation \cite{fang2022transferring,arnab2021vivit} for alignment. However, the neglect of local semantic extraction has hindered progress in applying large-scale models to the video retrieval domain \cite{ge2022miles, luo2020univilm}.

\begin{figure*}
    \begin{center}
        \includegraphics[width=1\linewidth]{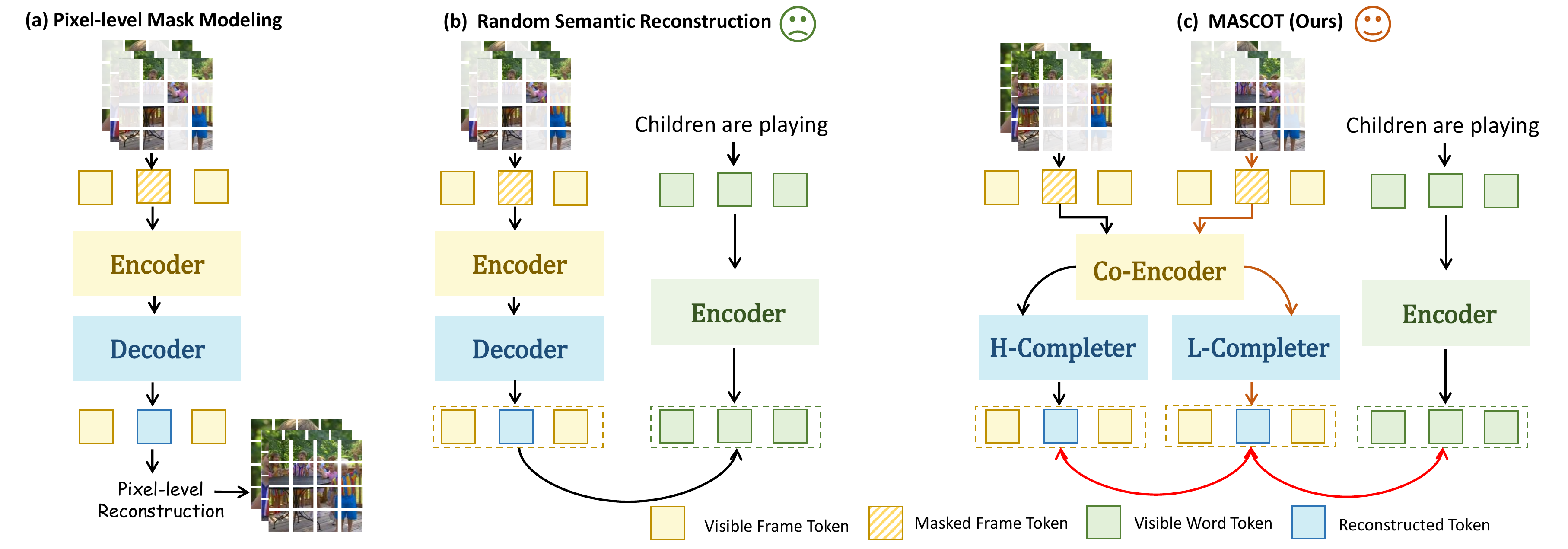}
    \end{center}
    \caption{(a) Mask for pixel-level reconstruction. (b) Employing random mask tubes for feature reconstruction. (c) Our MASCOT applies the low-informed and high-informed mask for prediction, prompting to understand the textual-aligned regions.
    }
    \label{fig:teaser}  
\end{figure*}

Masked video modeling (MVM) \cite{feichtenhofer2022masked,tong2022videomae} involves masking random tubes \cite{wang2023videomae,arnab2021vivit} and reconstructing the missing regions for pixel-level reconstruction (Fig. \ref{fig:teaser}\textcolor{red}{(a)}). To facilitate the comprehension of local semantics by the Vision Transformer (ViT) \cite{RN710}, a substantial proportion of patches in the tube are masked to reduce temporal redundancy. As depicted in Fig. \ref{fig:teaser}\textcolor{red}{(b)}, several methods \cite{hou2022milan,xue2021probing,shu2022masked,ma2022simvtp} employ MVM in video-text learning by adopting similar random masking strategies and using textual output as the reconstructed target to facilitate high-level perception extraction. However, determining which regions to mask for masked video prediction and learning text-related local content in the video remains challenging. We argue that random masking is inefficient in capturing text-related content. Since the masking is random, the unmasked parts \cite{kakogeorgiou2022hide} can still reveal the identity of visual concepts and aligned textual content. Furthermore, using the same reconstruction strategy to handle different masked regions also hinders model learning.

To address these limitations, we propose mask for semantics completion, called \textbf{MASCOT}, which utilizes a dual-completer to enhance local text-aligned correlations for masked semantics prediction. By leveraging self-attention in ViT, we can discern intrinsic relationships and generate masks for video tubes based on the attention map. The tokens that receive the most attention emphasize both video information and regions that are semantically similar to the text. Consequently, we introduce attention-based video masking to create different masks, with high-informed masks concealing more discriminative cues and low-informed masks disregarding areas containing irrelevant textual content.

To fully capture local content related to the text in videos, we employ video and text encoders to construct unmasked video and textual representations for building masked video prediction targets. 
As shown in Fig. \ref{fig:teaser}\textcolor{red}{(c)}, we use the spatial encoder as a co-encoder and feed the masked videos to model masked visual representations. 
Then, the proposed dual-completer is employed to remove noise, by reasoning corrupted details from  the correlation of neighboring visible patches. 
Specifically, unidirectional interaction is incorporated into the attention mechanism of the co-encoder and H-completer. The masked tokens can interact with unmasked tokens, leveraging all visible tokens to recover text-related semantics, while the attention interaction of unmasked tokens is unidirectionally neglected for de-noising. To disregard attention irrelevant to the text with low-informed masks, we introduce a shift of the reconstructed prediction with the L-completer from unaligned backgrounds to discriminative parts. Furthermore, we unify both two types of completions. A novel co-learning strategy is presented to provide more text-related attention distribution and integrate video cues.

In summary, our work contributes in four main ways:
\textbf{(1)} We propose a new perspective of assembling video-language learning  and MVM to align with text-related patches in the video.
\textbf{(2)} We introduce a novel video masking strategy that integrates textual content  to generate high-informed and low-informed masks.
\textbf{(3)} We propose a novel co-learning strategy to incorporate masking and reconstructed strategies with informed semantics completion and background attention shift. 
\textbf{(4)} We conduct extensive experiments and show that our MASCOT achieves new records on  MSR-VTT (54.8\%), LSMDC (28.0\%), ActivityNet (53.9\%), and DiDeMo (52.3\%).

\begin{figure*}
    \begin{center}
        \includegraphics[width=1\linewidth]{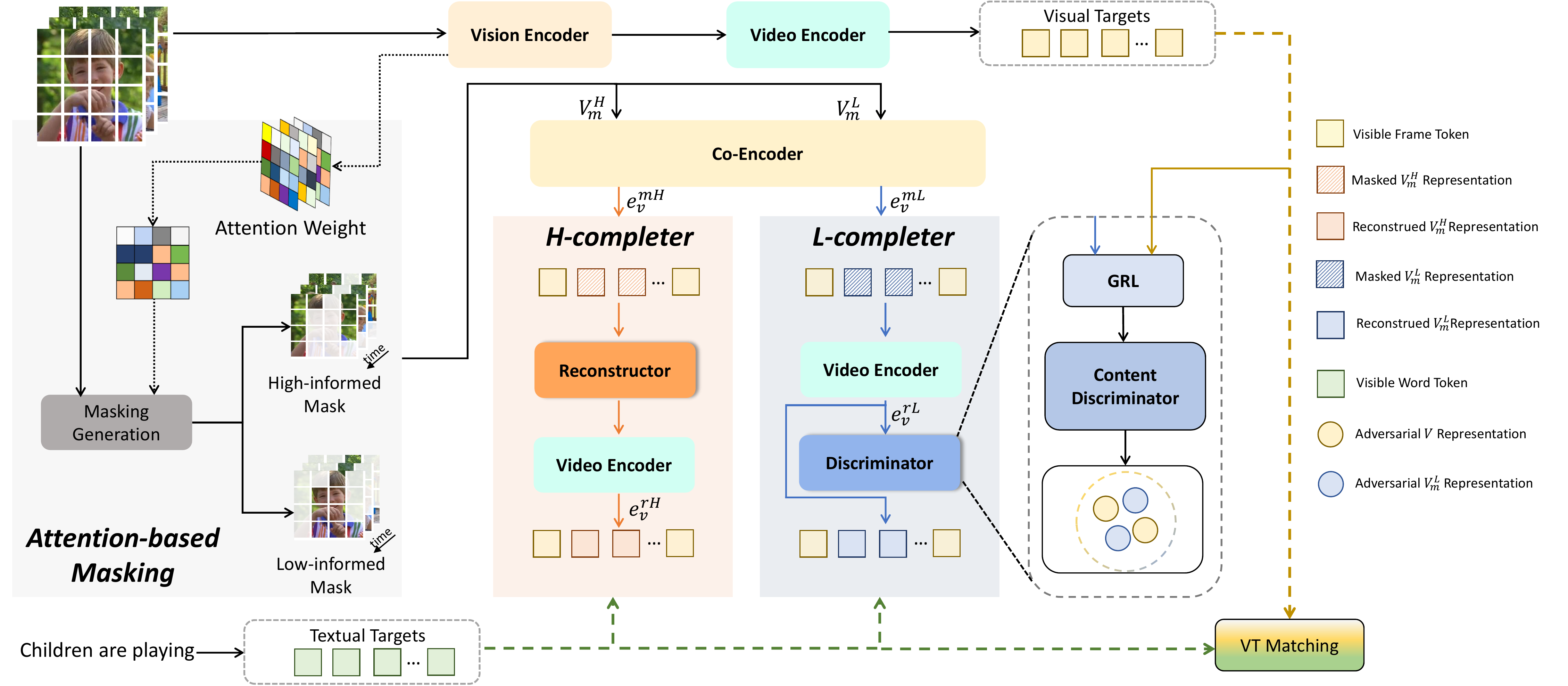}
    \end{center}
    \caption{Overview of MASCOT.  With the attention-based masking strategy, co-learning of the dual-completer is proposed, including Informed Semantics Reconstruction and Background Attention Shift. GRL represents gradient reversal layer \cite{ganin2015unsupervised}.
    }
    \label{framework}  
\end{figure*}

\section{Related Works}

\textbf{Video-text Retrieval.}
Existing methods for video-text retrieval \cite{dzabraev2021mdmmt,faghri2017vse++,liu2019use} utilize multiple representations from different modalities as input and employ fusion networks \cite{yu2018joint,wang2021t2vlad}. The video representation is integrated to align with the text content extracted by text encoders, such as BERT \cite{devlin2018bert}. Recently, researchers have focused on designing intensive cross-modal alignment mechanisms by incorporating pre-trained CLIP \cite{RN26} for video-language alignment. Since CLIP contains generalized content, CLIP4Clip \cite{luo2022clip4clip}, CLIP2video \cite{fang2022transferring}, and STAN \cite{liu2023revisiting} transfer the cross-modal correlations to the video level, employing various temporal modeling blocks. DRL \cite{wang2022disentangled}, X-CLIP \cite{ma2022x}, and X-pool \cite{gorti2022x} propose token-level interaction to adaptively exploit pair-wise correlations. Moreover, Ts2Net \cite{liu2022ts2}, CenterCLIP \cite{zhao2022centerclip}, and CLIP-ViP \cite{xue2022clip} incorporate spatial-temporal patch contexts into the vision encoder. In contrast to these methods, we propose masking text-related video tubes and reconstructing them to encourage the model to capture fine-grained semantic relations at both spatial and temporal levels.

\textbf{Masked Vision Modeling.}
 Masked visual encoders have been proposed to introduce masked visual modeling for self-supervised learning. Masked Image Modeling (MIM) \cite{he2022masked} masks a certain proportion of random 2D patches and optimizes the encoder and decoder for pixel-level reconstruction. BEiT \cite{bao2021beit} converts the image into a discrete sequence of tokens for masked token prediction, rather than directly recovering pixels. MVP \cite{wei2022mvp} introduces multi-modality guided teacher models in MIM, utilizing CLIP \cite{RN26} features as the reconstruction target. Moreover, masked video modeling \cite{feichtenhofer2022masked, tan2021vimpac, tong2022videomae} has been widely exploited by extending 2D masks into the video domain. VideoMAE \cite{tong2022videomae} proposes a tube mask strategy and achieves promising performance with a high mask ratio. MVD \cite{wang2022masked} suggests a two-stage masked feature modeling framework, employing knowledge distillation for the student model to predict high-level features from teacher models. In contrast, our MASCOT masks video patches for semantic completion and focuses on reconstructing salient regions, without relying on random masking.


\textbf{Masked Modeling for Cross-modal Alignment.}
Masked video modeling has been preliminarily explored in video-text alignment for feature-level completion. MILES \cite{ge2022miles} employs a snapshot video encoder updated by EMA \cite{he2020momentum} to produce reconstruction targets for handling randomly masked tubes. VidLP \cite{shu2022masked} introduces a simple framework for masking both video and text tokens, reducing spatial and temporal redundancy to enhance pre-training efficiency. In contrast to these methods, which adopt random masking and apply the same reconstruction strategy for various masked videos, our MASCOT generates high-informed and low-informed masks to intentionally mask text-related patches. Furthermore, we propose a dual-completer to address corrupted videos under different masks and capture more cross-modal content at the patch level.

\section{Method}


\subsection{Overview}
As illustrated in Fig. \ref{framework}, we adopt a two-stream encoder to encode video and text separately. 
We uniformly sample $M$ frames as video clips and feed unmasked video tokens into the vision transformer, along with temporal transformers \cite{luo2022clip4clip}, to model spatial and temporal information, generating video targets $e_v = \{v_0, v_1, \cdots, v_{M-1}\}$. 
In text encoding, each text is appended with two special tokens, \texttt{<SOS>} and \texttt{<EOS>}, and fed into the text encoder to embed textual representations $e_t = \{t_{\texttt{SOS}}, t_1, \cdots, t_{N-2}, t_{\texttt{EOS}}\}$ as textual targets, with $N$ denoting the text token number. Besides, we employ weighted token interaction (WTI) \cite{wang2022disentangled} to calculate the token similarity, fully exploiting the pair-wise token correlations and maximizing the similarity between positive pairs based on all tokens.

\subsection{Attention-based Video Masking}
Videos exhibit repetitive patterns in both spatial and temporal dimensions, allowing randomly masked areas to be easily restored by their neighboring regions. To enhance local semantic reasoning from unrelated patches, we propose attention-based tube masking, enabling the model to recover masked semantics by relating textual and unmasked visual information.
Specifically, we begin by creating a 2D mask for each frame. Each frame is divided into $n = hw/p^2$ non-overlapping patches, where $h$ and $w$ represent the image sizes, and $p$ denotes the patch size. The patches are then projected into $d$-dimensional tokens, with the $CLS$ token $Z^{CLS}$ prepended to extract the global representation, forming $Z \in \mathbf{R}^{(n+1)\times d}$. Then, multi-head self-attention (MSA) maps in the vision transformer are exploited:
\begin{equation}
\begin{split}
\label{MSA}
\mathcal{A}^l = \frac{1}{H}\sum_{i=1}^{H} softmax(Q^{l}_i K^{l\intercal}_i / \sqrt{d/H}),
\end{split}
\end{equation}
where $Q$ and $K$ represent the query and key transformed by tokens $Z^l$. $l$ denotes the $l$-th layer of MSA, and $H$ refers to the number of heads. 
In each layer, the correlation between each patch is modeled to provide the attention matrix averaged over all heads. To emphasize the relations between local regions and the global representation, we use the first row of the matrix (excluding the first element) as the attention weight: $w^l=\{w_{cls\cdot 1}, w_{cls\cdot 2}, \cdots, w_{cls\cdot n}\}$. To extend the attention weight to the 3D tube, we randomly sample two values $a_H^s$ and $a_H^e$ to indicate the start and end of the masked tube. The attention weights are then averaged over the masked tube, with each weight representing the relevance of the region concerning both the global video information and textual content.
To obscure more discriminative cues, we generate high-informed masks based on the attention weight's descending order. We choose a number $k=\lfloor r_H\cdot n \rfloor$, which is proportional to the number of patches with a mask ratio of $r_H \in [0, 1]$. We erase the top-$k$ patches by replacing them with random pixels and repeat them in the temporal dimension to obtain $V_{m}^H$, where the same spatial patches are masked for each frame in the masked video tube, leading to a more challenging masked video prediction. Similarly, we generate low-informed masks by sorting the weight in ascending order. The regions containing irrelevant textual content are masked with the masked spatial ratio of $r_L \in [0, 1]$ and masked temporal interval $a_L^s$ and $a_L^e$ to generate $V_{m}^L$.


    \begin{figure}
    \begin{center}
        \includegraphics[width=1\linewidth]{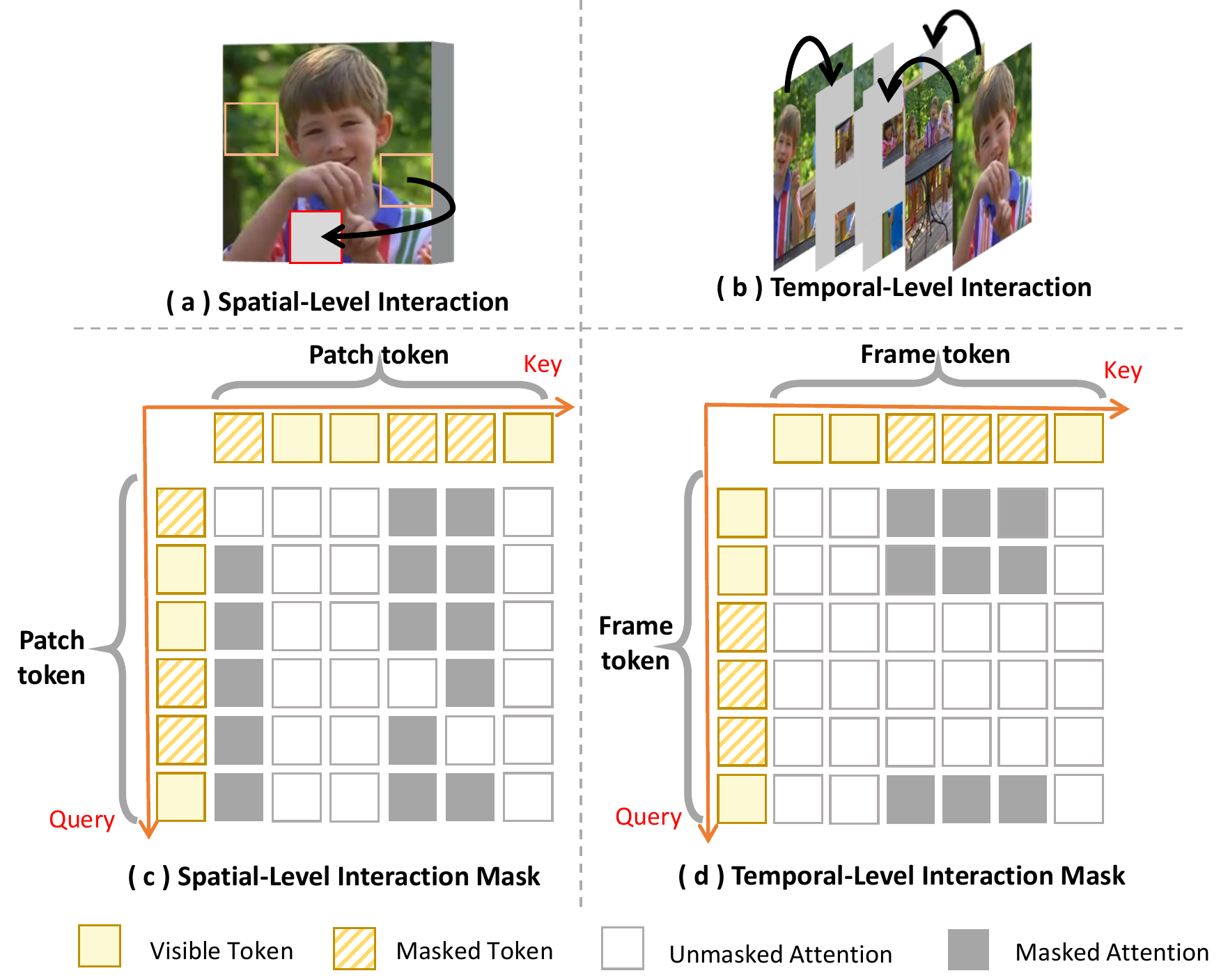}
    \end{center}
    \caption{Spatial patch and temporal tube attention interaction. The masked tokens can interact with unmasked tokens,  for unidirectional semantics completion, while the attention interaction of unmasked tokens is partially neglected.
    }
    \label{fig:high-informed}  
\end{figure}

\subsection{Informed Semantics Reconstruction}
Masked semantic completion encourages the model to focus on cross-modal details at the patch level by providing masked video prediction and reconstructing semantics through correlations between neighboring visible patches. In this section, we utilize the more challenging masked video $v_m^H$, where the masks conceal the majority of cross-modal content. To eliminate noise from masked salient regions, we propose a co-encoder and H-completer with a novel unidirectional interaction for recovering masked text-related semantics. Specifically, we follow a four-step process: \textbf{first}, the co-encoder (spatial encoder) is employed to model spatial information $e_v^{mH}$ and recover semantics at the spatial level. \textbf{Second}, a reconstructor is proposed for denoising encoding at the temporal level. \textbf{Third}, the reconstructed output is fed into the temporal transformer (Video Encoder) to obtain the reconstructed video representation $e_v^{rH}$ as the prediction. \textbf{Finally}, reconstruction is applied to minimize the semantic distance with visual and textual targets.

To fully explore the correlation of local patches and temporal tubes, we propose unidirectional interaction and impose it into every layer of co-encoder and reconstructor.
The attention calculation at $i$-th row and $j$-th column of $l$-th layer can be seen as follows:
\begin{equation}
\begin{split}
\label{MSA}
\mathcal{P}^l_{s}(i,j)/\mathcal{F}^l_{t} (i,j)= [(A^{l} \odot \mathcal{U}_{s/t}^{l}) \otimes V ^ {l}] (i, j) =   \sum_{k=1}^{K} a_{ik}^{l} u_{ik}^{l} v_{kj}^{l},
\end{split}
\end{equation}
where $\odot$ and $\otimes$ denote element-wise and matrix multiplication, $\mathcal{U}^{l}_{s/t}$ indicates spatial ($\mathcal{U}^{l}_{s}$) and temporal ($\mathcal{U}^{l}_{t}$) interaction mask, $\mathcal{P}^l_{s}$ is the attention output of co-encoder, and $\mathcal{F}^l_{t}$ is the attention output of reconstructor. $A$ and $V$ denote the attention weight and value respectively. In spatial encoding, all the attention calculations are the same as the original vision encoding in CLIP \cite{RN26}, except for adding spatial interaction masks. 
As depicted in Fig. \ref{fig:high-informed}\textcolor{red}{(c)}, the masked tokens adopted as the query, can attend all the unmasked neighboring patch tokens including global $cls$ token, while the unmasked tokens are only allowed to interact with visible information for de-noising. 
Specifically, we denote the patch tokens in co-encoders as $P = \{p_{cls}^{u}, p_{1}^{u}, p_{2}^{m}, p_{3}^{m} ..., p_{n}^{u}\} \in R^{(n+1) \times d}$, where $p^{m}$ represents the masked token and $p^{u}$ represents the unmasked token. The spatial interaction mask is calculated as:
\begin{equation}
\mathcal{U}_s(i,j) =
\left\{
\begin{aligned}
& 0 &&  p_{j} \in p^{m} (i\neq j),  \\
& 1 && \text{otherwise},
\end{aligned}
\right.  \label{encoder}
\end{equation} 
where the value of $0$ indicates that attention is neglected, $1$ represents the adoption of attention, and $i$ and $j$ represent the index of token $p$. The  representation of masked tokens has limited effects on unmasked regions. By unidirectional interaction, the model is trained to reconstruct masked regions using unmasked cues.


Before feeding the input into the video encoder for temporal modeling, we employ a single-layer self-attention layer as the reconstructor for completion at the temporal level. The reconstructor accepts the output of the co-encoder $e_{v}^{mH} = \{ v_{0}^{u}, v_{1}^{m}, v_{2}^{m},  ..., v_{M-1}^{u} \}$, where $v_{}^{u}$ and $v_{}^m$
denote the unmasked frame tokens and the frame tokens encoded by the masked portions of patches, respectively. We treat each frame token the same as the patch token  and impose temporal interaction masks into the reconstruction process.

As depicted in Fig. \ref{fig:high-informed}\textcolor{red}{(d)}, temporal interaction enables mutual attention among masked tokens, since the representation of masked frame tokens is partially recovered. This mask is formulated as:
 \begin{equation}
 \mathcal{U}_t(i,j) =
\left\{
\begin{aligned}
& 0 && v_{i} \in v^{u},  v_{j} \in v^{m}, \\
& 1 && \text{otherwise}.
\end{aligned}
\right.  \label{encoder}
\end{equation}
This completion enables that reconstructing video representation from high-informed masks, providing a more challenging prediction for capturing the fine-grained semantics at the patch level.

\subsection{Background Attention Shift}
To emphasize the modeling of salient regions, we introduce Background Attention Shift in conjunction with generated low-informed masks. The L-completer, which consists of the video encoder and content discriminator, is utilized to facilitate the shift of the reconstructed target from unaligned backgrounds to more discriminative parts. The video representation $e_v^{rL} = \{{v_0^L, v_1^L, \cdots, v_{M-1}^L}\}$ is generated by feeding $V_m^L$ into the co-encoder and the temporal transformer for low-informed video prediction. Unlike reconstructing representation from $V_m^H$, we only employ spatial reconstruction within the co-encoder. We conjecture  that there are minor distribution discrepancies between unmasked $e_v$ and masked $e_v^{rL}$ based on the following observations: 1. The masked regions, selected based on low attention weight, exhibit limited correlation with global $cls$ tokens. The videos, which lose irrelevant information, can still reveal the intrinsic identity; 2. The video obscured by low-informed masks can be easily recovered by applying unidirectional interaction from discriminative neighbors to text-irrelevant patches.

Therefore, we adopt adversarial learning to make low-informed video representation $e_v^{rL}$ indistinguishable from visual targets. The co-encoder and video encoder are optimized by maximizing the adversarial loss functions, while the content discriminator is trained in the opposite direction. 
Thus, this process can be formulated as:
 \begin{equation}
\begin{split}
\min_D \max_G \;\;&\mathcal{L}_{adv}(G,D)=\boldsymbol{E}_{(x,y)}\sum_{i=1}^\mathcal{D} \mathbf{1}[i=y] log(D(G(x))),
\end{split}
\end{equation}
where $y$ denotes the set of labels indicating masked and unmasked output, while $\mathcal{D}$ represents the number of domains. $G$ and $D$ correspond to the feature generation process and content discriminator. To simultaneously optimize $G$ and $D$, the gradient reversal layer (GRL) \cite{ganin2015unsupervised} is employed, reversing the gradient by multiplying it with a negative scalar during backward propagation. By employing adversarial learning, indistinguishable distributions between $e_v^{rL}$ and $e_v$ are obtained, encouraging the model to disregard the semantics of irrelevant backgrounds and shift the attention distribution, enhancing the correlation between modality-aligned regions.

\begin{figure}
    \centering
    \includegraphics[width=1\linewidth]{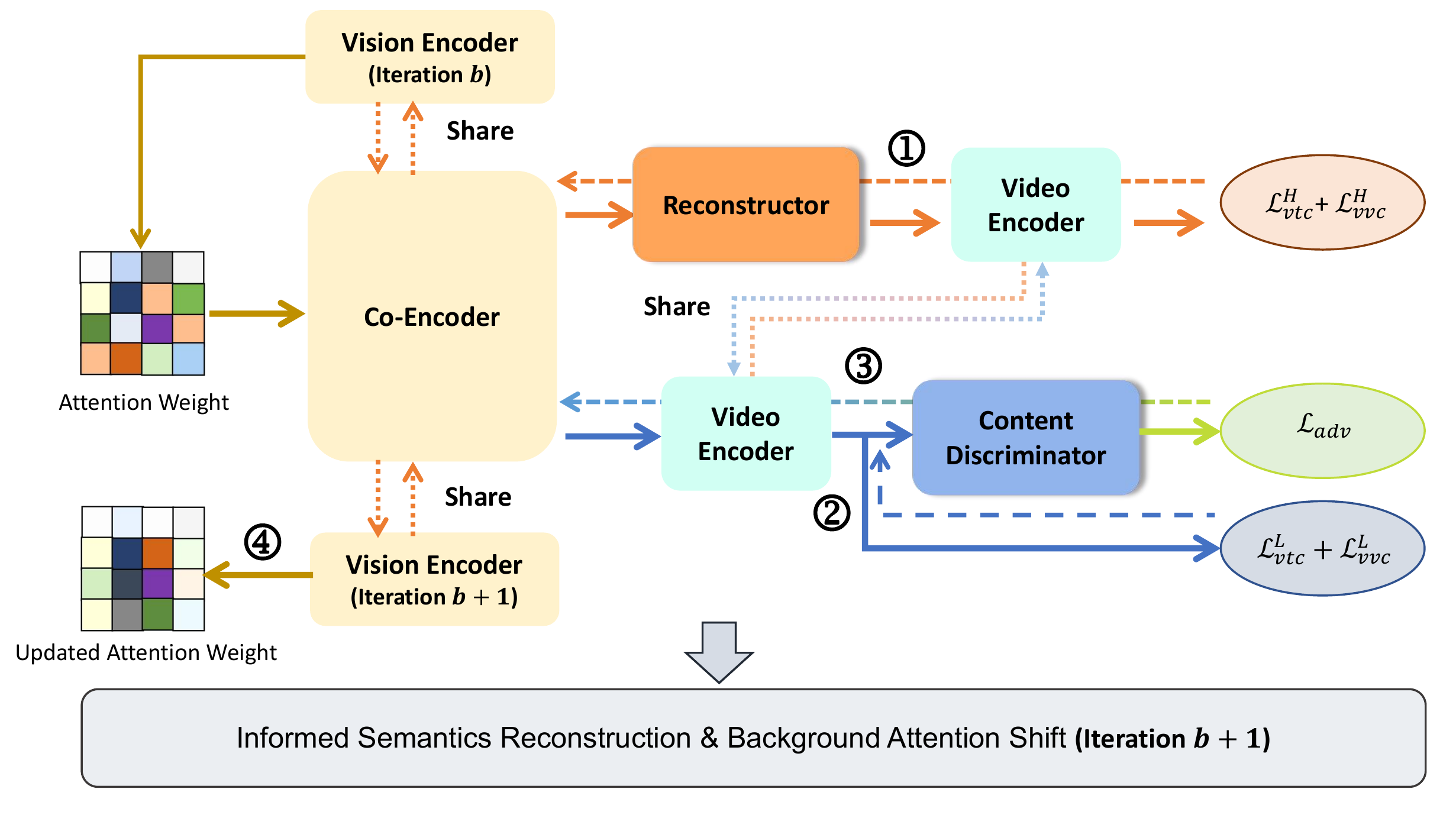}
    \caption{The four-step training scheme “Co-Learning”. The solid line and dashed line represents forward and back propagation.
    The dotted line indicates parameter sharing.}
    \label{fig:co-learning}
\end{figure}
\subsection{Dual-masking Co-learning}

To unify different completions and generate masks more relevant to cross-modal content for prediction, we design a collaborative learning strategy called co-learning, as illustrated in Fig. \ref{fig:co-learning}. Co-learning consists of four learning branches for each iteration.
\textbf{First}, the H-completer is updated by processing the masked video $v_m^H$ for informed semantics reconstruction.
\textbf{Second}, the video encoder within the L-completer is updated by processing the masked video $v_m^L$ for contrastive learning.
\textbf{Third}, the gradient of $\mathcal{L}_{adv}$ is used to update the content discriminator and reversed to enhance attention shift for the video and spatial encoder.
\textbf{Finally}, the spatial encoder, which shares parameters with the co-encoder, is also updated to provide a more fine-grained distribution for attention-based masking in the subsequent iteration.
Overall, co-learning allows the branches to mutually promote each other. By recovering high-informed and low-informed masked regions, the model is trained to concentrate on modeling text-aligned regions and discard attention to irrelevant backgrounds.
Consequently, the training objective of contrastive loss based on WTI \cite{wang2022disentangled} can be formulated as:
\begin{equation}
    \mathcal{L}_{x_1 2 x_2}=-\frac{1}{B} \sum_i^B \log \frac{\exp \left(\operatorname{WTI}\left(\mathbf{e}_{x_1, i}, \mathbf{e}_{x_2, i}\right) / \tau\right)}{\sum_j^B \exp \left(\operatorname{WTI}\left(\mathbf{e}_{x_1, i}, \mathbf{e}_{x_2, j}\right) / \tau\right)},
      \label{total}
\end{equation}
\begin{equation}
    \mathcal{L}_{x_2 2 x_1}=-\frac{1}{B} \sum_i^B \log \frac{\exp \left(\operatorname{WTI}\left(\mathbf{e}_{x_2, i}, \mathbf{e}_{x_1, i}\right) / \tau\right)}{\sum_j^B \exp \left(\operatorname{WTI}\left(\mathbf{e}_{x_2, i}, \mathbf{e}_{x_1, j}\right) / \tau\right)},
\end{equation}
\begin{equation}
    \mathcal{L}_{x_1x_2c}=\frac{1}{2}(\mathcal{L}_{x_1 2 x_2} + \mathcal{L}_{x_2 2 x_1}),
      \label{total}
\end{equation}
where $\mathbf{e}_{x_1}$ and $\mathbf{e}_{x_2}$ are the output tokens of different modalities and $\mathcal{L}_{x_1 2 x_2}$ is optimized to minimize the distance between two token sequences. The batch size is represented by $B$, and $\tau$ denotes the temperature.
In our method, we adopt both visual and textual targets as objectives and present training reconstruction targets as:
\begin{equation}
    \mathcal{L}_{total}  = \mathcal{L}_{vtc}  + \alpha(\mathcal{L}_{vtc}^{H} + \mathcal{L}_{vvc}^{H}) + \beta(\mathcal{L}_{vtc}^{L} + \mathcal{L}_{vvc}^{L}) + \gamma\mathcal{L}_{adv},
    \label{total}
\end{equation}
where $\mathcal{L}_{vtc}^{L}$ and $\mathcal{L}_{vtc}^{H}$ represent loss within pairs between masked video and textual target.
$\mathcal{L}_{vvc}^{L}$ and $\mathcal{L}_{vvc}^{H}$ is optimized by pulling masked video and visible video targets together in embedding space.

\section{Experiment}
\begin{figure}
    \centering
    \includegraphics[width=1\linewidth]{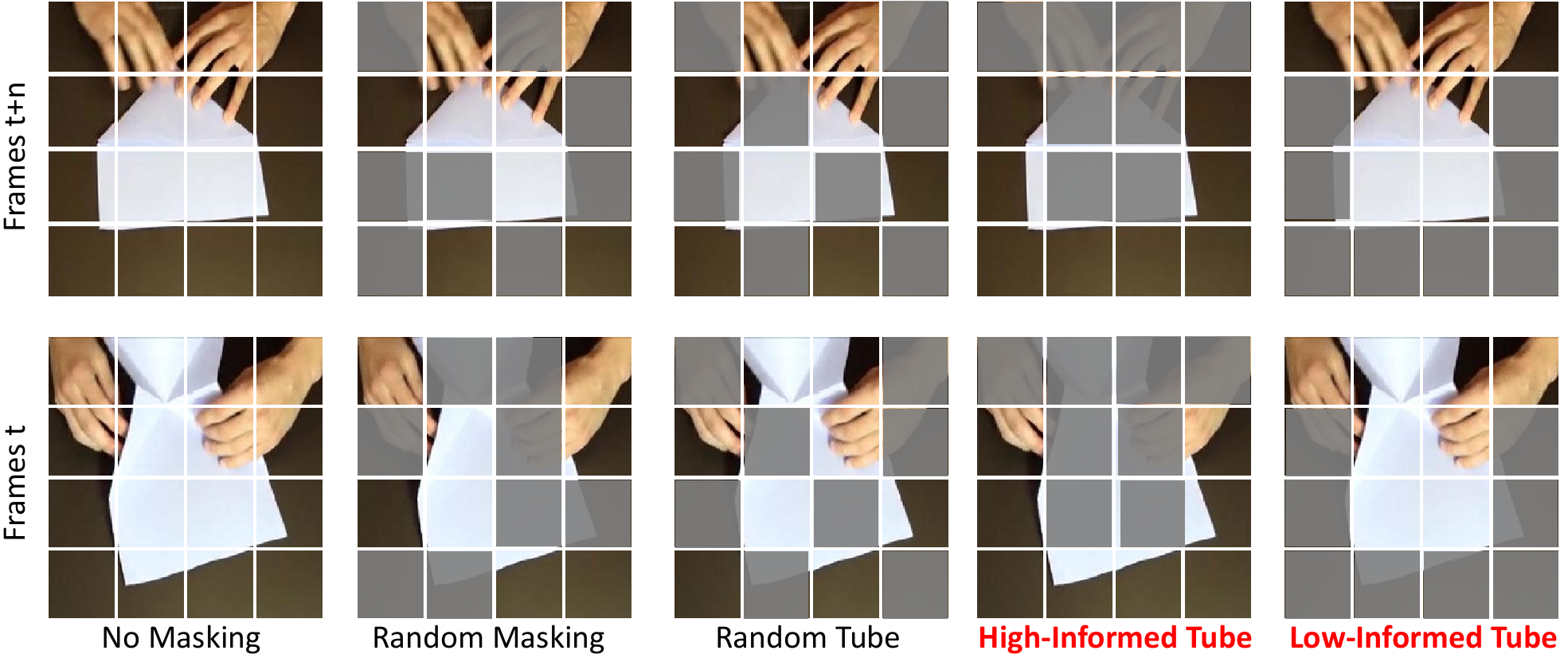}
    \caption{An illustration of different masking strategies.}
    \label{fig:mask}
\end{figure}

\subsection{Experimental Setting}
\textbf{Datasets.}
\emph{MSR-VTT} \cite{xu2016msr} comprises 10,000 videos, each of which contains 20 captions. We adopt two training splits, referred to as 7k \cite{miech2019howto100m} and 9k \cite{gabeur2020multi}, which consist of 7,000 and 9,000 videos, respectively. For evaluation, we use the 1K-A test set \cite{yu2018joint}, which includes 1,000 videos.
\emph{LSMDC} \cite{rohrbach2017movie} is made up of 118,081 videos sourced from 202 movies, with 1,000 videos selected for evaluation. \emph{DiDeMo} \cite{anne2017localizing} includes over 10,000 videos, of which 8,395 videos are used for training and 1,004 videos for testing. Following \cite{lei2021less,liu2021hit,luo2022clip4clip}, we concatenate all descriptions of a video into a single caption and treat this dataset as a video-paragraph retrieval task. \emph{ActivityNet Captions} \cite{krishna2017dense} comprises 20k videos with 100k dense descriptions. Following the video-paragraph retrieval setting \cite{liu2019use, zhang2018cross}, we train our models on 10k videos and test on 4.9k videos.

\textbf{Evaluation Metric.} Following standard video-text retrieval metrics \cite{miech2018learning,mithun2018learning,zhang2018cross}, we  report Recall at rank K (R@K), median rank (MdR) and mean rank (MnR) as metrics, where K=1, 5, 10 are calculated. The higher R@K, lower median rank and mean rank indicate better performance. Besides, we also sum up all the R@K results as Rsum \cite{liu2022ts2} to indicate the overall performance.

\textbf{Implementation Details.}
We employ the vision and text encoder, initialized by CLIP, to encode frame and word tokens. The backbone of MASCOT is ViT-B/32, with a patch size $p$ of 32. Then, we adopt a 4-layer temporal transformer with 8 heads and 512 channels, included in both the H-completer and L-completer. Two linear projections encode the different modalities into a 512-dimension space to match their representations. For MSR-VTT \cite{xu2016msr} and LSMDC \cite{rohrbach2017movie}, the video and caption lengths are set to 12 and 32, respectively. Following existing methods \cite{liu2019use, zhang2018cross}, we set both video and caption lengths to 64 for DiDeMo \cite{anne2017localizing} and ActivityNet \cite{krishna2017dense}. Our MASCOT is trained using the Adam optimizer with a cosine schedule decay and a batch size of 128. For MSR-VTT \cite{xu2016msr} and LSMDC \cite{rohrbach2017movie}, we train for 5 epochs, while for DiDeMo \cite{anne2017localizing} and ActivityNet \cite{krishna2017dense}, we train for 10 epochs. Besides, the learning rate is set to 1e-7 for the frame encoder and text encoder and 1e-4 for the proposed modules. We utilize the final MSA layer of ViT as attention weight and set the mask ratios $r_H$ and $r_L$ to 0.7 and 0.5, respectively, while the temporal intervals $a_L^s$, $a_L^e$, $a_H^s$, and $a_H^e$ are randomly set within the maximum video length. Additionally, we set $\alpha$, $\beta$, and $\gamma$ to 0.5, 0.5, and 0.2, respectively, to control the weight of loss functions.

\begin{table}
 		\caption{Ablation studies on the masking strategy. "Rand" is shortened for "random mask". “Tube” denotes that the masking the same patches along the temporal dimension. Text-video retrieval on MSR-VTT-9k \cite{gabeur2020multi} are evaluated.}
	\begin{center}
	\resizebox{0.48\textwidth}{!}{
		\begin{tabular}{c|c|ccc|ccc|c} 
	\hline
    \multicolumn{2}{c}{}&\multicolumn{3}{c}{Text $\Longrightarrow$  Video} &\multicolumn{3}{c}{Video $\Longrightarrow$ Text} \\
    \hline
	Masking&Ratio & R@1 & R@5 & R@10 & R@1 & R@5 & R@10 & Rsum \\
    \hline
-  &-&46.6         &	73.4&	83.5&	45.4&	73.4&	81.9 & 404.2 \\
Rand  &50\% &45.8  &	74.9&	83.3&	45.0 &	72.9&	83.8 & 405.7 \\
Rand  &70\% &46.2  &	75.5&	83.6&	44.6 &	73.4&	83.7 & 407.0 \\
Rand + Tube &50\%  &47.6&	73.3&	83.3&	45.3&	75.2&	83.3 & 408.0 \\
Rand + Tube &70\%  &47.1&	75.4&	84.3&	44.5&	74.7&	83.2 & 409.2 \\
Low-informed &50\% &47.9&	75.5&	83.7&	46.0&	74.3&	84.4 & 411.8 \\
Low-informed &70\% &47.3&	74.7&	83.5&	45.2&	73.6&	83.9 & 408.2 \\
High-informed &50\% & 47.8 & 75.5 & 84.1 & 46.2 & 74.5 & 83.3 & 411.4 \\
High-informed &70\% &48.0 & 75.7 & 84.0 & 46.2 &  74.7 & 84.5 & 413.1\\ 
\hline
	\end{tabular}}
 \end{center}
	\label{mask_strategty}
\end{table} 

\begin{table}
 		\caption{Effects of different components, where the text-video retrieval results are evaluated on MSR-VTT-9k \cite{gabeur2020multi}. "$V$" and "$T$" represents the co-encoder and video encoder. "$D$" denotes the discriminator.  "$R$" denotes the reconstructor. "$V_{u}$" and "$R_u$" indicates that co-encoder and reconstructor with unidirectional interaction. "$H+L$" represents co-learning. }
	\begin{center}
	\resizebox{0.48\textwidth}{!}{
		\begin{tabular}{c|ccc|ccc|c} 
	\hline
    \multicolumn{1}{c}{}&\multicolumn{3}{c}{Text $\Longrightarrow$  Video} &\multicolumn{3}{c}{Video $\Longrightarrow$ Text} \\
    \hline
Method  & R@1 & R@5 & R@10 & R@1 & R@5 & R@10 & Rsum \\
    \hline
    \rowcolor{gray!10}\multicolumn{2}{c}{}&\multicolumn{3}{c}{\textit{H-completer}}& \multicolumn{2}{c}{}&\\
    \hline
$T$ + $V$             &47.1 & 75.0 & 83.2&	45.5 &	73.9 &	83.2 & 407.9 \\ 
$T$ +$V$ +  $R$       &47.5 & 75.5 & 83.6&	46.2 &	74.5 &	83.8 & 411.1\\ 
$T$ + $V$ +  $R_u$    &47.8 & 75.7 & 83.5&	46.1 &	74.6 &	84.0 & 411.7\\ 
$T$ +$V_u$ +   $R$    &47.9 & 75.5 & 83.8&	46.0 &	74.7 &	84.3 & 412.2\\ 
$T$ + $V_u$ +  $R_u$ ($H$)  &48.0 & 75.7 & 84.0 & 46.2 &  74.7 & 84.5 & 413.1\\ 
\hline
\rowcolor{gray!10}\multicolumn{2}{c}{}&\multicolumn{3}{c}{\textit{L-completer}}& \multicolumn{2}{c}{}&\\
\hline
$T$ +$V$          &47.6&	75.0&	83.4&	45.7&	73.4&	83.4 & 408.5\\
$T$ +$V$ + $R_u$  &47.2&	74.4&	83.6&	45.0&	73.1&	83.5 & 406.8\\
$T$ +$V_u$        &47.5&	75.1&	83.5&	45.6&	73.7&	83.9 & 409.3\\
$T$ + $V$ + $D$  &47.4&    75.3&	83.7&	45.8&	74.2&	84.0 & 410.4\\
$T$ +$V_u$ + $D$ ($L$)  &47.9&	75.5&	83.7&	46.0&	74.3&	84.4 & 411.8\\
\hline
\rowcolor{gray!10}\multicolumn{2}{c}{}&\multicolumn{3}{c}{\textit{Co-learning}}& \multicolumn{2}{c}{}&\\
\hline
$H + L$  &48.3 &75.8 &84.4  & 46.6&	74.4& 85.0 & \textbf{414.5} \\
\hline
	\end{tabular}}
 \end{center}
	\label{ablation}
\end{table}

\subsection{Ablation Experiments}






\textbf{Video masking strategy. }  
The diverse video masking strategies are investigated, with visualizations displayed in Fig. \ref{fig:mask}.
Random masking independently conceals random patches for each frame, while tube-based masking obscures the same blocks along the temporal dimension.
In Tab. \ref{mask_strategty}, by employing informed semantics completion for reconstruction, random masks yield the worst results.
Using tube-based masking for the model to reason masked semantics solely from neighboring patches proves more effective.
Additionally, high-informed masks deliberately conceal regions containing salient cross-modal content, pushing the model to temporally and spatially integrate visible patches for fine-grained reconstruction at the patch level.
Conversely, low-informed masking for background attention shift demonstrates less satisfactory performance, indicating that reasoning the masked text-related patches is essential.
We also investigate the impact of the mask ratio and observe that both high-informed and low-informed masks perform well at ratios of 70\% and 50\%, respectively. It is found that creating a more difficult high-informed reconstruction task improves model learning. However, masking too many text-irrelevant regions for attention shift may result in the neglect of crucial semantic information.

\textbf{Effects of H-completer.}
We explore various structures of H-completer for informed semantic completion, including the reconstructor, and spatial and temporal interaction mask. 
As depicted in the upper part of Tab. \ref{ablation}, incorporating a reconstructor between the co-encoder and video encoder yields a substantial improvement. This result indicates that temporal relations act as crucial cues for predicting masked video patches. 
Consequently, the model enhances its reasoning ability regarding local spatial correlation at the temporal level, while the video encoder can be focused on temporal modeling.
Furthermore, we introduce unidirectional interaction and apply it to both the co-encoder and reconstructor.  The masked tokens are adopted as query to interact with unmasked tokens in spatial interaction and all the features in temporal interaction for completion, leveraging all visible video cues.
Meanwhile, the attention interactions of visible correlation are only allowed between the unmasked tokens for unidirectional de-noising. Compared to full attention ($T +V + R$), the inclusion of unidirectional interaction ($T + V_u + R_u$) results in further performance enhancement. Similar improvement is observed in the middle of Tab. \ref{ablation}, where spatial interaction ($T + V_u$) is imposed for background attention shift.



\textbf{Effects of L-completer.} It is worth noting that incorporating a reconstructor in the L-completer yields inferior results compared to solely adopting a "spatial-video" structure. 
This can be attributed to the fact that focusing on reconstructing text-irrelevant regions does not enhance the semantic correlation at the patch level, leading to decreased performance. 
As videos masked by low-informed masks still reveal sufficient intrinsic identity for cross-modal alignment, we employ the content discriminator $D$ to minimize the representation difference. The model is encouraged to learn indistinguishable feature distributions between unmasked and masked representations without reconstruction. 
Consequently, the masked video representation, obscured by regions with low attention weight, demonstrates a similar performance to unmasked videos by mitigating the effects of patch masking. This suggests that the model identifies and disregards more text-irrelevant information, achieving attention shifting towards regions with more cross-modal content.

 \begin{figure}
    \centering
    \includegraphics[width=1\linewidth]{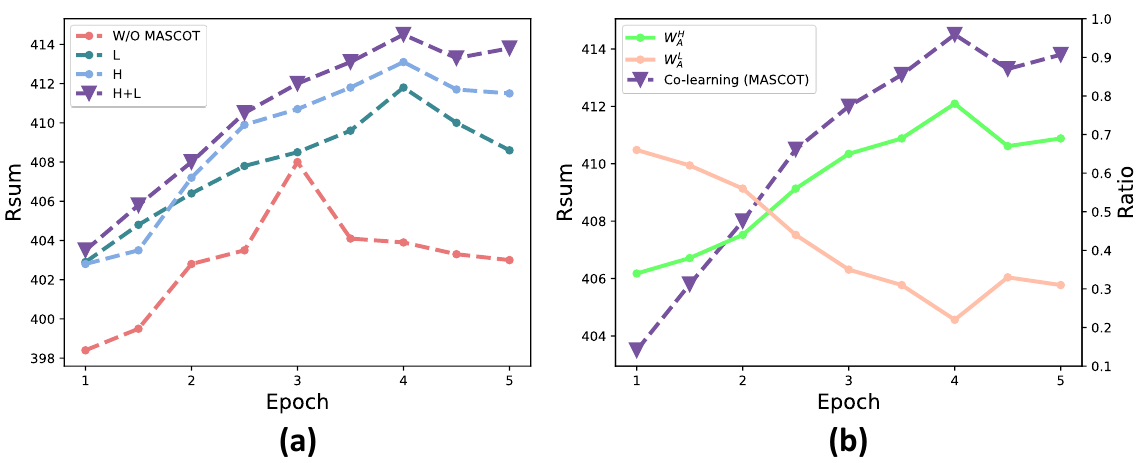}
    \caption{\textbf{(a)} Analysis of Rsum with co-learning. "W/O MASCOT" represents without reconstruction. \textbf{(b)} Attention weight versus performance in MSR-VTT-9k \cite{xu2016msr}. $W_A^{H}$ and $W_A^{L}$ represent the top and bottom 30$\%$ of attention weights.}
    \label{fig:acc}
\end{figure}

\begin{table*}
\centering
\caption{Retrieval performance comparison trained on MSR-VTT-9k \cite{xu2016msr} and evaluated on 1k-A test. * means adopting DSL \cite{cheng2021improving} as extra trick during inference.}
		\resizebox{0.8\textwidth}{!}{
			\begin{tabular}{c|ccccc|ccccc} 
    \hline
	Method & R@1 & R@5 & R@10 & MdR & MnR & R@1 & R@5 & R@10 & MdR & MnR\\
    \hline
    CE \cite{liu2019use} & 20.9 & 48.8 & 62.4 & 6.0 & 28.2 & 20.6 & 50.3 & 64.0 & 5.3 & 25.1 \\
	MMT \cite{gabeur2020multi}    & 26.6 & 57.1 & 69.6	& 4.0  & 24.0 & 27.0 & 57.5 & 69.7 & 3.7 & 21.3 \\
	SUPPORT-SET \cite{patrick2020support}    & 27.4 & 56.3 & 67.7 & 3.0  & -    & 26.6 & 55.1 & 67.5 & 3.0 & - \\
	T2VLAD \cite{wang2021t2vlad}             & 29.5 & 59.0 & 70.1 & 4.0  & -    & 31.8 & 60.0 & 71.1 & 3.0 & -  \\
	FROZEN \cite{bain2021frozen}            & 31.0 & 59.5 & 70.5	& 3.0  & -    & -    &    - & -    &   - & - \\
    HIT \cite{liu2021hit}         & 30.7 & 60.9 & 73.2 & 2.6  & -    & 32.1 & 62.7 & 74.1 & 3.0 & - \\
	MDMMT \cite{dzabraev2021mdmmt}           & 38.9 & 69.0 & 79.7	& 2.0  & 16.5 & -    &    - & -    &   - & - \\
    \hline
\rowcolor{gray!10}\multicolumn{1}{c}{\textit{CLIP-ViT-B/32}}& & & & & & & & & & \\
    \hline
	CLIPforward \cite{portillo2021straightforward} & 31.2 & 53.7 & 64.2 & 4.0  & -    & 27.2 & 51.7 & 62.6 & 5.0 & - \\
    CLIP4Clip \cite{luo2022clip4clip} & 44.5 & 71.4 & 81.6	& 2.0  & 15.3 & 42.7 & 70.9 & 80.6 & 2.0 & 11.6 \\
    CLIP2Video \cite{fang2022transferring} & 45.6 & 72.6 & 81.7	& 2.0  & 14.6 & 43.5 & 72.3 & 82.1 & 2.0 & 10.2 \\
    QB-Norm \cite{bogolin2022cross} &47.2 &73.0 &83.0 &2.0 & - & - & - & - & -& - \\
    STAN \cite{liu2023revisiting} & 46.9 & 72.8 & 82.8 & 2.0 & - & - & - & - & - & - \\
    STAN* \cite{liu2023revisiting} & 49.0 & 74.8 & 83.5 & 2.0 & - & - & - & - & - & - \\
    CAMOE* \cite{cheng2021improving} &47.3 &74.2 &84.5 &2.0 & 11.9 & 49.1 & 74.3 & 84.3 & 2.0& 9.9 \\
    X-pool \cite{gorti2022x}  & 46.9 & 72.8 & 82.2	& 2.0  & 14.3 & - & - & - & - & - \\
    X-CLIP \cite{ma2022x} & 46.1 & 73.0 & 83.1 & 2.0 & 13.2 & 46.8 & 73.3 & 84.0 & 2.0 & 9.1\\
    TS2Net \cite{liu2022ts2} & 47.0 & 74.5 & 83.8	& 2.0  & 13.0 & 45.3 & 74.1 & 83.7 & 2.0 & 9.2 \\
    DRL \cite{wang2022disentangled} &47.4 &74.6 &83.8 &2.0 &- & 45.3 & 73.9 & 83.3 & 2.0 & 8.7 \\
        \hline
    \rowcolor{gray!30}MASCOT (Ours) & \textbf{48.3} & \textbf{75.8} & \textbf{84.4}	& \textbf{2.0}  & \textbf{12.6} & \textbf{46.6} & \textbf{74.4} & \textbf{85.0} & \textbf{2.0} & \textbf{9.1} \\
    \rowcolor{gray!30}MASCOT* (Ours) & \textbf{52.5} & \textbf{78.3} & \textbf{85.9}	& \textbf{1.0}  & \textbf{10.5} & \textbf{52.8} & \textbf{78.2} & \textbf{86.3} & \textbf{1.0} & \textbf{8.6} \\
	\hline
\rowcolor{gray!10}\multicolumn{1}{c}{\textit{CLIP-ViT-B/16}}& & & & & & & & & & \\
    \hline
    CLIP2TV \cite{gao2021clip2tv}& 48.3 & 74.6 & 82.8 & 2.0 & 14.9 & 46.5 & 75.4 & 84.9 & 2.0 & 10.2 \\
    CenterCLIP  \cite{zhao2022centerclip} & 48.4 & 73.8 & 82.0 & 2.0 & 13.8 & 47.7 & 75.0 & 83.3 & 2.0 & 10.2 \\
	TS2Net \cite{liu2022ts2} & 49.4 & 75.6 & 85.3	& 2.0  & 13.5 & 46.6 & 75.9 & 84.9 & 2.0 & 8.9 \\
    X-CLIP \cite{ma2022x} & 49.3 & 75.8 & 84.8 & 2.0 & 13.2 & 48.9 & 76.8 & 84.5 & 2.0 & 8.1 \\
    DRL \cite{wang2022disentangled} & 50.2 & 76.5 & 84.7 & 1.0 & - & 48.9 & 76.3 & 85.4& 2.0 & - \\
    STAN \cite{liu2023revisiting} & 50.0 & 75.2 & 84.1 & 1.5 & - & - & - & - & - & - \\
    STAN* \cite{liu2023revisiting} & 54.1 & 79.5 & 87.8 & 1.0 & - & - & - & - & - & - \\
    \hline
	\rowcolor{gray!30}MASCOT (Ours)  & \textbf{50.5} & \textbf{77.6} & \textbf{85.2} & \textbf{1.0}  & \textbf{11.2}  & \textbf{49.5} & \textbf{77.3} & \textbf{86.4} & \textbf{2.0} & \textbf{8.0} \\
 \rowcolor{gray!30}MASCOT* (Ours)  & \textbf{54.8} & \textbf{80.3} & \textbf{87.7} & \textbf{1.0}  & \textbf{9.6}  & \textbf{54.9} & \textbf{79.4} & \textbf{87.6} & \textbf{1.0} & \textbf{7.3} \\
	\hline
		\end{tabular}}
	\label{MSRVTTTable}
\end{table*}

\begin{table}
	\caption{Text-to-video results on MSR-VTT-7k \cite{miech2019howto100m}. * means adopting DSL \cite{cheng2021improving} as extra trick during inference.}
	\begin{center}
		\resizebox{0.45\textwidth}{!}{
			\begin{tabular}{c|ccccc} 
    \hline
	Method & R@1 & R@5 & R@10 & MdR & MnR \\
    \hline
ClipBERT \cite{lei2021less}          & 22.0 & 46.8 & 59.9	  & 6.0  & -    \\
CLIP4Clip \cite{luo2022clip4clip}      & 42.1 & 71.9 & 81.4 & 2.0 & 16.2 \\
X-Pool  \cite{gorti2022x}  & 43.9 & 72.5 & 82.3 & 2.0 & 14.6  \\
\hline
\rowcolor{gray!30}\textbf{MASCOT (Ours)} & \textbf{45.8} & \textbf{73.0} & \textbf{83.1} & \textbf{2.0} & \textbf{13.9} \\
\rowcolor{gray!30}\textbf{MASCOT* (Ours)} & \textbf{47.2} & \textbf{75.8} & \textbf{85.3} & \textbf{2.0} & \textbf{12.4} \\
\hline
\end{tabular}}
	\end{center}
 
	\label{mSrvtt7k}
\end{table}

\begin{table}
\caption{Text-to-video results on LSMDC \cite{rohrbach2017movie} testing set. * means adopting DSL \cite{cheng2021improving} as extra trick during inference.}
	\begin{center}
		\resizebox{0.45\textwidth}{!}{
			\begin{tabular}{c|ccccc} 
    \hline
	Method & R@1 & R@5 & R@10 & MdR & MnR \\
    \hline
MDMMT \cite{dzabraev2021mdmmt}  & 18.8 & 38.5 &47.9 &12.3 & 58.0    \\
CLIP4Clip \cite{luo2022clip4clip}   & 22.6 & 41.0 & 49.1 & 11.0 & 61.0 \\
STAN \cite{liu2023revisiting} & 23.7 & 42.7 & 51.8 & 9.0 & - \\
DRL \cite{wang2022disentangled} & 24.9 & 45.7 & 55.3 & 7.0&- \\
STAN* \cite{liu2023revisiting} & 26.2 & 46.0 & 53.9 & 9.0 & - \\
CAMOE* \cite{cheng2021improving}&25.9&46.1&53.7 & - & 54.4 \\
\hline
\rowcolor{gray!30}\textbf{MASCOT (Ours)} & \textbf{25.9} & \textbf{47.5} & \textbf{58.0} & \textbf{7.0} & \textbf{44.7} \\
\rowcolor{gray!30}\textbf{MASCOT* (Ours)} & \textbf{28.0} & \textbf{48.1} & \textbf{58.8} & \textbf{6.0} & \textbf{43.3} \\
\hline
		\end{tabular}}
	\end{center}	
	\label{LSMDC}
\end{table}

\begin{table}
		\caption{Text-to-video results on DiDeMo \cite{anne2017localizing} testing set. * means adopting DSL \cite{cheng2021improving} as extra trick during inference.}
	\begin{center}
		\resizebox{0.45\textwidth}{!}{
			\begin{tabular}{c|ccccc} 
\hline
Method & R@1 & R@5 & R@10 & MdR & MnR \\
\hline
TeachText \cite{croitoru2021teachtext} & 21.1 & 47.3 & 61.1 & 6.3 & -  \\
CLIP4Clip \cite{luo2022clip4clip}      & 43.4 & 70.2 & 80.6 & 2.0 & 17.5 \\
TS2-Net \cite{liu2022ts2}              & 41.8 & 71.6 & 82.0 & 2.0 & 14.8   \\
STAN \cite{liu2023revisiting} & 46.2 & 70.4 & 80.0 & 2.0 & - \\
STAN* \cite{liu2023revisiting} & 51.3 & 75.1 & 83.4 & 1.0 & - \\
\hline
\rowcolor{gray!30}\textbf{MASCOT (Ours)} & \textbf{48.1} & \textbf{75.7} & \textbf{85.1} & \textbf{2.0} & \textbf{11.8} \\
\rowcolor{gray!30}\textbf{MASCOT* (Ours)} & \textbf{52.0} & \textbf{78.3} & \textbf{85.3} & \textbf{1.0} & \textbf{10.0} \\
\hline
\end{tabular}}
\end{center}
	\label{DiDemo}
\end{table}

\begin{table}
		\caption{Text-to-video results on ActivityNet \cite{krishna2017dense} testing set. * means adopting DSL \cite{cheng2021improving} as extra trick during inference.}
	\begin{center}
		\resizebox{0.45\textwidth}{!}{
			\begin{tabular}{c|ccccc} 
    \hline
Method & R@1 & R@5 & R@10 & MdR & MnR \\
\hline
ClipBERT \cite{lei2021less} & 21.3 & 49.0& 63.5 & 6.0 & - \\
MMT \cite{gabeur2020multi}           & 28.7 & 61.4 & -	  & 3.3  & 16.0    \\
CLIP4Clip \cite{luo2022clip4clip}      & 40.5 & 72.4 & - & 2.0 & 7.5 \\
TS2-Net \cite{liu2022ts2}              & 41.0 & 73.6 & 84.5 & 2.0 & 8.4   \\
\hline
\rowcolor{gray!30}\textbf{MASCOT (Ours)} & \textbf{45.4} & \textbf{76.1} & \textbf{87.2} & \textbf{2.0} & \textbf{5.8} \\
\rowcolor{gray!30}\textbf{MASCOT* (Ours)} & \textbf{53.9} & \textbf{81.0} & \textbf{89.9} & \textbf{1.0} & \textbf{4.9} \\
\hline
		\end{tabular}}
	\end{center}
	\label{ActivityNet}
\end{table}

 \begin{figure*}[t]
    \centering
    \includegraphics[width=1\linewidth]{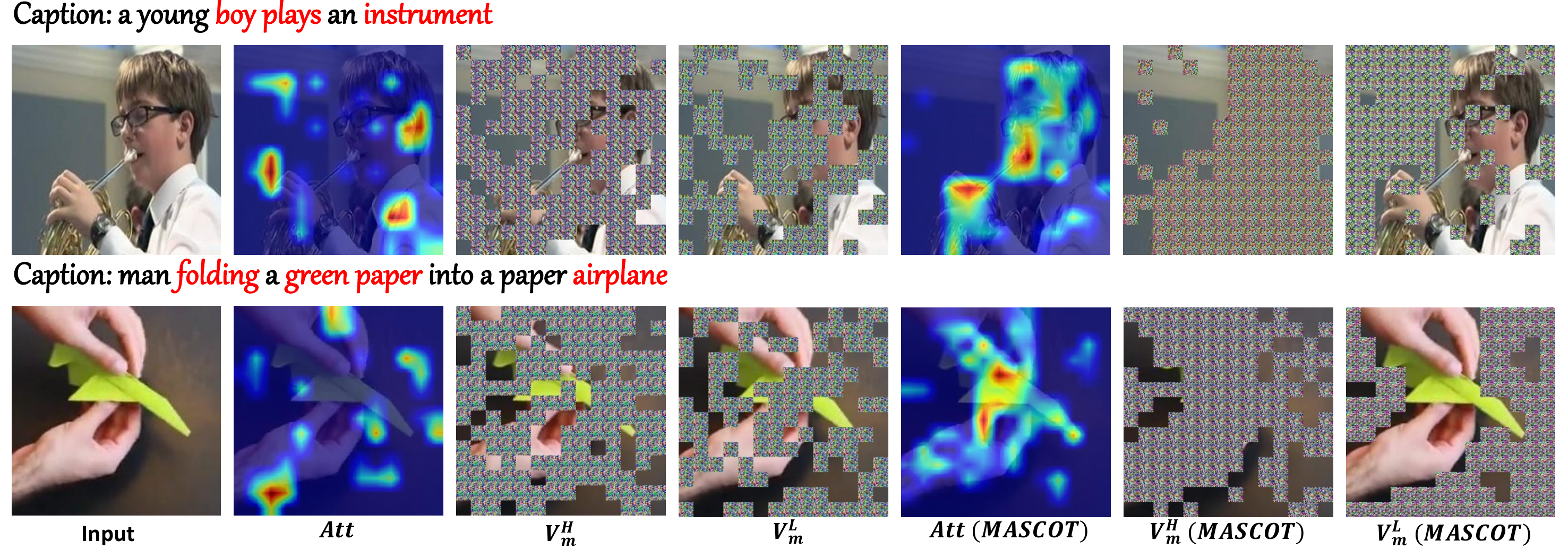}
    \caption{$Att$, $V_m^H$, and $V_m^L$ represent the attention map, video masked by high-informed and low-informed mask, where the results in columns 2-4 and columns 5-7 are initialized by CLIP \cite{RN26} and obtained after training with MASCOT, respectively.
    }
    \label{fig:vis1}
\end{figure*}
\textbf{Effects of Co-learning.}
To demonstrate the effectiveness of incorporating both H-completer and L-completer, we present the results of co-learning ($H+L$) at the bottom of Tab. \ref{ablation}, which achieves significant improvement. Co-learning enables the two completions to complement each other by discarding semantics of irrelevant backgrounds and focusing more on completing the masked semantics of cross-modal content. In Fig. \ref{fig:acc}\textcolor{red}{(a)}, we illustrate the performance curves across different epochs. The performance without MASCOT initially increases when transferring image representations into videos but rapidly decreases afterward, indicating that the model suffers from overfitting due to limited training data. However, by gradually adding the H-completer and L-completer, Rsum improves and eventually stabilizes, which signifies the robustness of mitigating overfitting by introducing challenging masked video prediction. 
Furthermore, we visualize the change of average attention weight for attention-based masking in Fig. \ref{fig:acc}\textcolor{red}{(b)}. The top $30\%$ ($W_A^{H}$) and bottom $30\%$ ($W_A^{L}$) attention weights are depicted, which are used to generate high-informed and low-informed masks. As performance increases, the top and bottom attention weights increase and decrease, respectively. This indicates that the model's learning focus has indeed shifted from irrelevant patches to regions containing more cross-modal content, leading to better performance.

\subsection{Comparisons with State-of-the-art Models} 

To fairly evaluate the performance of our proposed MASCOT, we compare it with other state-of-the-art methods on four datasets: MSR-VTT \cite{xu2016msr}, LSMDC \cite{rohrbach2017movie}, DiDeMo \cite{anne2017localizing}, and ActivityNet \cite{krishna2017dense}. The text-to-video results are presented in Tab. \ref{MSRVTTTable}, \ref{mSrvtt7k}, \ref{LSMDC}, \ref{DiDemo}, and \ref{ActivityNet}. As observed, our method outperforms the previous methods without any extra tricks by a significant margin, improving R@1 by at least \textbf{3.0\%} across all four benchmarks. Specifically, we achieve \textbf{+3.8\%} and \textbf{+3.3\%} higher R@1 than CLIP4Clip \cite{luo2022clip4clip} with ViT-B/32 on MSR-VTT-9k \cite{xu2016msr} and LSMDC \cite{rohrbach2017movie}.
Compared to the leading method TS2-Net \cite{liu2022ts2}, significant improvements can be observed on DiDeMo \cite{anne2017localizing} (\textbf{+6.3\%} at R@1) and ActivityNet \cite{krishna2017dense} (\textbf{+4.4\%} at R@1), where longer video tubes are masked to recover masked prediction and identify text-related semantics. Additionally, the results of MSRVTT-9k achieved by CLIP-B/16 are reported in Tab. \ref{MSRVTTTable}, which still outperforms the previous best method. Furthermore, we include results with inverted softmax (DSL \cite{cheng2021improving}), which significantly improves the performance. This indicates that our model learns the generalized video-language alignment at the fine-grained level. The prominent
results demonstrate that MASCOT works well on cross-modal pairs. 

\begin{figure}
    \centering
    \includegraphics[width=0.92\linewidth]{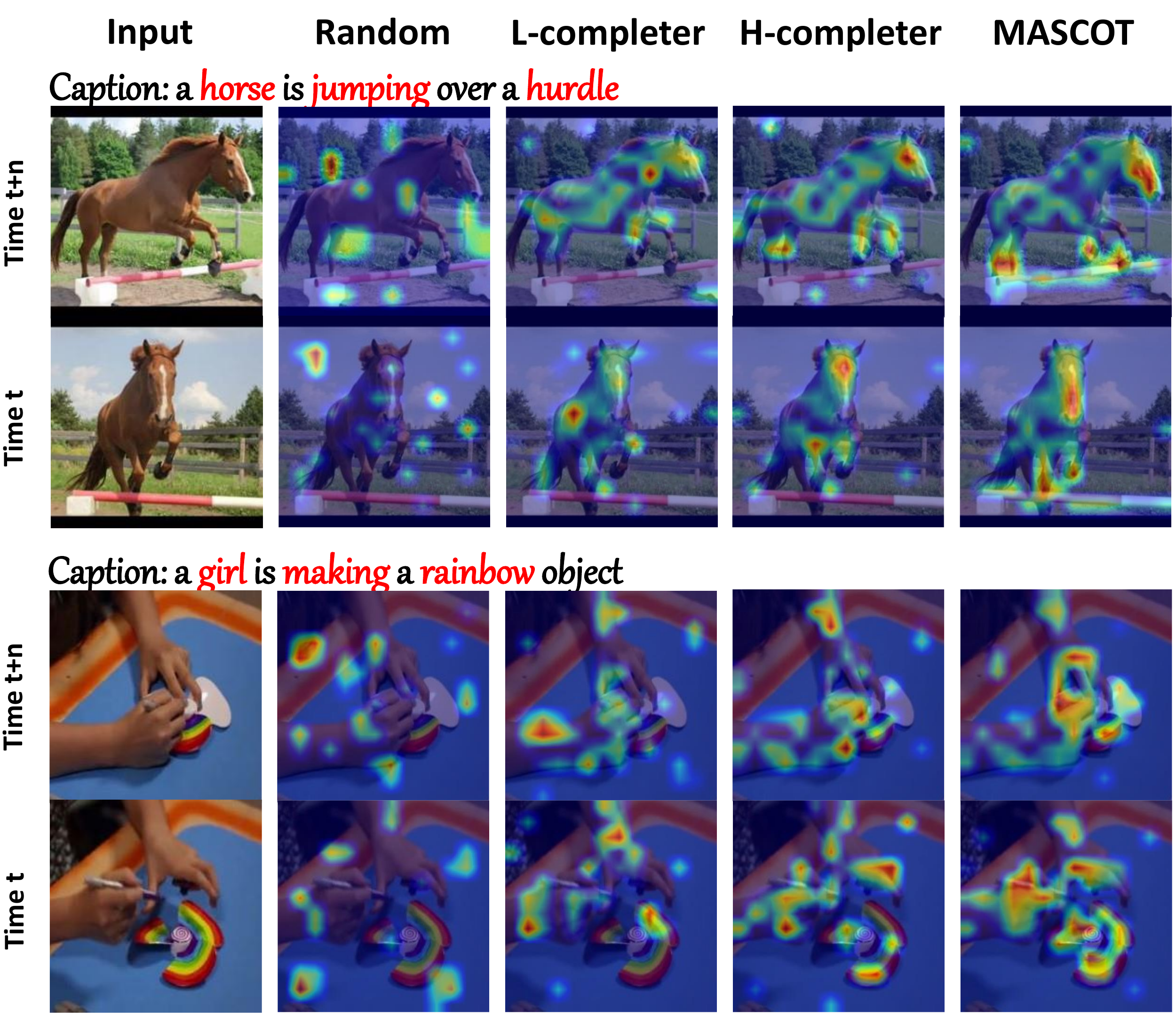}
    \caption{The visualizations of attention weight in the co-encoder. "Random" means adopting random masks for reconstruction. "L-completer" and "H-completer" are based on low and high-informed masks, respectively, while MASCOT unifies both two masks and corresponding reconstruction strategies to capture text-aligned visual semantics.}
    \label{fig:vis2}
\end{figure}

\subsection{Qualitative Results}

We present a visualization of the self-attention map obtained from the final layer of the co-encoder, based on ViT-B/16 architecture in Fig. \ref{fig:vis1}. This map is used to generate both high-informed and low-informed masks. To demonstrate the effectiveness of MASCOT's masking and reconstruction strategies, we show the changes in the attention map before and after training. Our results show that MASCOT is successful in shifting attention from irrelevant patches to relevant video cues, resulting in the capture of more cross-modal correspondences.
Besides, we also provide illustrations of change in the masked frames, where the pixel-level masks are depicted to intentionally obscure semantic-based regions.
This indicates that attention-based masking gradually captures more fine cross-modal content for challenging reconstruction with training.

We also present Fig. \ref{fig:vis2}, which visualizes the attention distribution trained using different strategies. Our proposed MASCOT model pays more attention to local regions with text-related content compared to the model with random masks. 
Specifically, in the first two rows, MASCOT focuses on the hurdle and the shape of a horse, including the movement of horse hooves, while the model with random masks identifies regions of limited content, such as the body and some background, which are not relevant to the caption. By utilizing attention-based masking and incorporating the H-completer and L-completer, MASCOT captures cross-modal content at the local level while neglecting attention to irrelevant regions.

\section{Conclusion}

We propose \textbf{MA}sk for \textbf{S}emantics \textbf{CO}mple\textbf{T}ion, a novel method for video-language learning and masked video modeling. Our method employs attention-based masking to intentionally obscure videos by utilizing high-informed masks for informed semantics reconstruction and low-informed masks for background attention shift. Furthermore, our co-learning strategy incorporates two completions and enhances the fine-grained understanding of video context based on cross-modal content. The proposed method delivers significant improvements across four video-text benchmarks.




{\small
\bibliographystyle{ieee_fullname}
\bibliography{egbib}
}

\end{document}